\documentclass[acmsmall]{acmart}

\usepackage{url}
\usepackage{xcolor}
\definecolor{newcolor}{rgb}{.8,.349,.1}

\usepackage{graphicx}
\usepackage{booktabs}
\usepackage{xcolor,gensymb}
\usepackage{hyperref}
\usepackage{mathpazo}
\usepackage{amsmath}
\usepackage{longtable}
\usepackage{xcolor}
\usepackage[normalem]{ulem}

\usepackage{graphicx} 
\usepackage{float}
\usepackage{svg}
\usepackage{breakurl}

\def\BibTeX{{\rm B\kern-.05em{\sc i\kern-.025em b}\kern-.08emT\kern-.1667em\lower.7ex\hbox{E}\kern-.125emX}}
\definecolor{lightgray}{gray}{0.9}

\renewcommand{\textrightarrow}{$\rightarrow$}

\copyrightyear{2018}
\acmYear{2018}
\setcopyright{acmlicensed}
\acmConference[Woodstock '18]{Woodstock '18: ACM Symposium on Neural Gaze Detection}{June 03--05, 2018}{Woodstock, NY}
\acmBooktitle{Woodstock '18: ACM Symposium on Neural Gaze Detection, June 03--05, 2018, Woodstock, NY}
\acmPrice{15.00}
\acmDOI{10.1145/1122445.1122456}
\acmISBN{978-1-4503-9999-9/18/06}

\begin{document}

%
\title{Avoiding Overfitting: A Survey on Regularization Methods for Convolutional Neural Networks}

%
\author{Claudio Filipi Gon\c{c}alves dos Santos}
\email{cfsantos@ufscar.br}
\affiliation{%
  \institution{Federal Institute of S\~{a}o Carlos - UFSCar}
  \streetaddress{Rod. Washington Luiz, 235}
  \city{S\~{a}o Carlos}
  \state{S\~{a}o Paulo}
  \country{Brazil}}
\affiliation{%
  \institution{Eldorado's Institute of Technology}
  \streetaddress{Av. Alan Turing, 275}
  \city{Campinas}
  \state{S\~{a}o Paulo}
  \country{Brazil}}

\author{Jo\~{a}o Paulo Papa}
\email{joao.papa@unesp.br}
\orcid{0000-0002-6494-7514}
\affiliation{%
  \institution{S\~{a}o Paulo State University - UNESP}
  \streetaddress{Av. Eng. Lu\'{i}s Edmundo Carrijo Coube, 14-01}
  \city{Bauru}
  \state{S\~{a}o Paulo}
  \country{Brazil}
}

%
\renewcommand{\shortauthors}{Santos, et al.}
\newcommand{\claudio}[1]{\textcolor{blue}{#1}}
\newcommand{\papa}[1]{\textcolor{orange}{#1}}
\newcommand{\review}[1]{\textcolor{black}{#1}}

%
\begin{abstract}
Several image processing tasks, such as image classification and object detection, have been significantly improved using Convolutional Neural Networks (CNN). Like ResNet and EfficientNet, many architectures have achieved outstanding results in at least one dataset by the time of their creation. A critical factor in training concerns the network's regularization, which prevents the structure from overfitting. This work analyzes several regularization methods developed in the last few years, showing significant improvements for different CNN models. The works are classified into three main areas: the first one is called ``data augmentation", where all the techniques focus on performing changes in the input data. The second, named ``internal changes", which aims to describe procedures to modify the feature maps generated by the neural network or the kernels. The last one, called ``label", concerns transforming the labels of a given input. This work presents two main differences comparing to other available surveys about regularization: (i) the first concerns the papers gathered in the manuscript, which are not older than five years, and (ii) the second distinction is about reproducibility, i.e., all works refered here have their code available in public repositories or they have been directly implemented in some framework, such as TensorFlow or Torch.
\end{abstract}

%
%


%

%

%
\maketitle

\section{Introduction}
\label{s.introduction}

Convolutional neural networks (CNNs) have achieved relevant results on several computer vision-related tasks, such as image classification and object detection in scenes. Such success can be explained by how the convolutional neuron works: it highlights given features according to the spatial properties of the image. The initial layers highlight less complex features, such as borders; however, more dept layers can detect more complex traits, like entire objects or faces of people. Nowadays, it is hard to find any other computer vision technique applied without any CNNs, from biometrics to disease detection.

One key aspect concerning CNNs is how to stack the convolutional kernels to accomplish the best result on a given task. It is widespread to use the same basic architecture on several different tasks, just changing the output. For instance, the basic block used for EfficientNet~\cite{efficientnet}, a neural network used for image classification, is also used on the Efficient-Det~\cite{efficientdet} architecture to tackle the object detection task.

The architecture may be the central part of a computer vision model; however, there are other relevant points before starting the training step. For instance, the optimization technique can influence the final result. Even the kernels' initial random values can influence how well the model will perform in the end. This study focuses on one of these aspects that can influence the final result: the regularization algorithms. Depending on the chosen regularization strategy used, some architectures can achieve a relevant gain on the final results. One important aspect of using a good regularizer is that it does not influence the final model's performance. It means that, independently of using or not one regularizer, the model's computational cost for inference is the same. However, in some cases, it can influence performance during the training phase, using a little computational overhead or pre-train epochs. In any way, the results of the output usually overcompensate this cost. 

\subsection{How regularization works}
\label{ss.reg-works}

CNNs are usually used for computer vision tasks, such as image classification and object detection, to create models as powerful as human vision. If the amount of information available is considered, it becomes clear the training task requires more data variability than possible. Considering a healthy human with a regular brain and eyes, we retain new information around $16$ hours per day, on average, disregarding the time we sleep. Even considering huge datasets such as ImageNet, the number of images available is minimal compared to the quantity of data a human brain receives through the eyes. This unavailability of new data may lead to a situation known as overfitting, where the model learns how to represent well the training data, but it does not perform well on new information, i.e., the test data. This situation usually happens when the model has been trained exhaustively in the available training information that it cannot generalize well in other new information.

As an artificial neural network, the training step of CNNs can be described as an optimization problem, where the objective is to find out the weight values which, given an input and a loss function, can transform the information in the desired output, such as a label, with the lowest possible error. One way to achieve this goal is to minimize the following function:

\begin{equation}
\label{e.standard_optimization}
\min_{U, V} ||X - WY^T||_F^2,
\end{equation}
where $||.||_F^2$ is the Frobenius norm, $X\in\mathbb{R}^{m \times n}$ defines the input data, and $W\in\mathbb{R}^{m \times d}$ and $Y\in\mathbb{R}^{n \times d}$ denote the weight matrix and the target labels, respectively. According to~\cite{cavazza2018dropout}, the Frobenius norm imposes the similarity between $X$ and $WY^T$. This interpretation has one main advantage: this formulation enables the optimization through matrix factorization, producing a structured factorization of $X$. However, it is only possible to achieve a global minimum if $W$ or $Y^T$ is fixed for optimizing both matrices together converts the original equation into a non-convex formulation. This problem can be solved if the matrix factorization is changed to a matrix approximation as follows: 

\begin{equation}
\label{e.approximation_matrix}
\min_{A} ||X - A||_F^2,
\end{equation}
where the target is to estimate the matrix $A$, which ends up in a convex optimization, meaning it has a global minimum that can be found via gradient descent algorithms. When using regularization, this equation becomes:

\begin{equation}
\label{e.regularization}
\min_{A} ||X - A||_F^2  +  \lambda\Omega(A),
\end{equation}
where $\Omega(\cdot)$ describes the regularization function based on $A$, and $\lambda$ is the scalar factor that sets how much influence the regularization function infers on the objective function.

One key aspect of the regularization methods, independent of the training phase it works, is to prevent the model from overfitting the training data. It operates by increasing the variability of the data on different stages of a CNN. When working with images, the most straightforward method is random image changing, like rotation and flipping. Several deep learning frameworks, such as Keras and TensorFlow, have their implementation available, facilitating this kind of regularization and improving the results. Although this type of regularization works well, some points should be taken into consideration. For example, some transformations may distort the image into another existing class in the classification. The more straightforward example is baseline image classification on the MNIST data set: if the rotation is too several, an input "$6$" may be transformed into a "$9$", leading the model to learn wrong information.

\subsection{Regularization vs. Normalization}
\label{ss.regxnorm}

A general problem in machine learning is to tune the parameters of a given model to perform well on the training data and eventually new information, i.e., the test set. The collection of algorithms that aims to reduce the error on the data that does not belong to the training set is called regularization techniques. 

One main difference between the normalization and regularization techniques is that the second is not performed after the training period, while the first is kept in the model. For example, Cutout~\cite{cutout} and MaxDropout~\cite{maxdropout} original codes show they do not execute anything during the inference, but the BatchNormalization~\cite{batchnorm} executes its algorithm in deducing the test set.

\subsection{Scope of this work}
\label{ss.scope}

This study focuses on the most recent regularization techniques for CNNs. Other studies~\cite{surveydataaugmentation,surveydeepmodels} focus on older and more general regularization methods. Here, we consider three main points:

\begin{itemize}
\item \textit{Recently developed}: besides Dropout~\cite{dropout}, no other study is older than four years, making this study very much up-to-date;
\item \textit{Code availability}: all related algorithms in this study are available in some way, usually on Github. We considered it an essential point because it avoids studies with possibly inaccurate results and allows reproducibility when necessary;
\item \textit{Results}: all techniques here were able to improve the results of the original models significantly.
\end{itemize} 

In this work, the regularization algorithms are divided into three main categories, each one in a given section: the first one is called ``data augmentation", and it describes the techniques that change the input of a given CNN. The second category is called ``internal changes", and it describes the set of algorithms that changes values of a neural network internally, such as kernel values or weights. The third category is called ``label", in which techniques perform their changes over the desired output. Table~\ref{tbl:methods} gives a list of all methods discussed in this work.

\begin{table*}[htbp]
\centering

\begin{tabular}{lcll}
\multicolumn{1}{c}{Reference} & Short Name      & \multicolumn{1}{c}{Description}                                                                                                            & \multicolumn{1}{c}{Where} \\
\cite{bag}                             & Bag of Tricks          & \begin{tabular}[c]{@{}l@{}}Combines several regularizers\\ to show how it improves CNN\end{tabular}                                     & Input                     \\
\cite{batchaugment}                             & Batch Augment          & \begin{tabular}[c]{@{}l@{}}Increases the size of the\\ mini-batch\end{tabular}                                     & Input                     \\
\cite{fixres}                             & FixRes          & \begin{tabular}[c]{@{}l@{}}Performs train and test with different\\ image sizes\end{tabular}                                     & Input                     \\

\cite{cutout}                             & Cutout          & Removes part of the image                                                                                                                  & Input                     \\
\cite{cutmix}                             & CutMix          & \begin{tabular}[c]{@{}l@{}}Replaces part of the image using\\ other parts of other images\end{tabular}                                     & Input / Label                    \\
\cite{randomerasing}                             & RandomErasing   & \begin{tabular}[c]{@{}l@{}}Replaces part of the image by noise\\ or paint the region\end{tabular}                                          & Input                     \\
\cite{mixup}                             & Mixup           & \begin{tabular}[c]{@{}l@{}}Mixes two images from \\ different classes\end{tabular}                                                         & Input/Label               \\
\cite{autoaugment}                             & AutoAugment     & \begin{tabular}[c]{@{}l@{}}Learns how to provide better data\\ augmentation based on information\\ from the training data set\end{tabular} & Input                     \\

\cite{fastautoaugment}                             & Fast AutoAugment     & \begin{tabular}[c]{@{}l@{}}Reduces the training time of the agent \\from \cite{autoaugment} \end{tabular} & Input                     \\

\cite{randaugment}                             & RandAugment    & \begin{tabular}[c]{@{}l@{}}Learns augmentation policies \\during training \end{tabular} & Input                     \\
\cite{pba}                             & PBA     & \begin{tabular}[c]{@{}l@{}}Population Based algorithm \\ for data augmentation \end{tabular} & Input                     \\

\cite{cutblur}                             & CutBlur         & \begin{tabular}[c]{@{}l@{}}Replaces regions from high-resolution\\ images with low resolution pieces\end{tabular}                          & Input / Label              \\
\cite{dropout}                             & Dropout         & Drops random neurons                                                                                                                       & Internal                  \\
\cite{maxdropout}                             & MaxDropout      & \begin{tabular}[c]{@{}l@{}}Drops neurons based on their\\ activation\end{tabular}                                                          & Internal                  \\
\cite{gradaug}                             & GradAug  & \begin{tabular}[c]{@{}l@{}}Trains sub-networks from the original\\ CNN\end{tabular}                                                          & Internal                  \\
\cite{localdrop}                             & Local Drop      & \begin{tabular}[c]{@{}l@{}}Dropout and DropBlock based on the\\Radamacher complexity\end{tabular}                                                          & Internal                  \\

\cite{shakeshake}                             & Shake-Shake     & \begin{tabular}[c]{@{}l@{}}Gives different weights to each branch\\  of the residual connection\end{tabular}                               & Internal                  \\
\cite{shakedrop}                            & ShakeDrop       & \begin{tabular}[c]{@{}l@{}}Improves Shake-Shake by \\ generalizing to other models\end{tabular}                                            & Internal                  \\
\cite{manifoldmixup}                            & Manifold Mixup       & \begin{tabular}[c]{@{}l@{}}Act like Mixup, however,\\ in the middle layers of a CNN\end{tabular}                                            & Internal / Label                  \\
\cite{dropblock}                            & DropBlock       & Drops entire regions from a tensor                                                                                                         & Internal                  \\
\cite{autodropout}                            & AutoDrop       & Learns drop pattern & Internal                  \\
\cite{inceptionv2v3}                            & Label Smoothing & \begin{tabular}[c]{@{}l@{}}Replaces one-hot encoding vectors\\ to smoothed labels\end{tabular}                                             & Label    \\
\cite{tsla}                            & TSLA & \begin{tabular}[c]{@{}l@{}}Two-stage algorithm\\ for label smoothing\end{tabular}                                             & Label     \\
\cite{sls}                            & SLS & \begin{tabular}[c]{@{}l@{}}Quantifies label smoothing\\ based on feature space\end{tabular}                                             & Label     \\
\cite{jocor}                            & JoCoR & \begin{tabular}[c]{@{}l@{}}Co-relates labels\\ for label smoothing\end{tabular}                                             & Label               
\end{tabular}
\caption{Summarization of the approaches considered in the survey.}
\label{tbl:methods}
\end{table*}

\review{Although we divided the methods into three different strategies, Table~\ref{tbl:methods} highlights that some algorithms work on two different levels. For instance, CutMix and CutBlur work on both input and label levels. The majority of the methods work on input or internal structures, which shows a lack of research on label regularization methods.}


\subsection{Comparison with Other Works}
\label{ss.comparison}

In a quick search, it is possible to find a diversity of works using Convolutional Neural Networks, such as image classification~\cite{efficientnet,resnet,tinynet,squeeze}, object detection~\cite{r-cnn,yolo}, and image reconstruction~\cite{rdn-reconstruction, rdn-sr,dncnn}. However, the frequency of works for regularization compared to other problems is very low. As far as we are concerned, we found only two recent surveys about regularization for deep neural networks.

The first one~\cite{surveyreg0} is an extensive analysis of regularization methods and their results. Although it is an interesting work, it focuses considerably on older methods, such as adding noise to the input, DropConnect~\cite{dropconnect} and Bagging~\cite{bagging}. Those methods are still broadly used and have their importance; however, they are not exactly new. 

Another relevant work found was a survey focused only on dropout-based approaches~\cite{surveydropout}. Dropout~\cite{dropout} is undoubtedly an important regularization method for different types of neural networks, and it has influenced several new approaches over the years, besides being used in several different architectures. 

In this work, we show very recent developments on strategies for improving the results of Convolutional Neural Networks. As one can observe, it presents works as recent as the one published on 2021~\cite{localdrop,autodropout}. \review{The following sub-sections present more insights and statistical information about the works surveyed in the manuscript.}

\subsection{Where do regularizers work primarily?}
\label{ss.where_work}

\begin{figure}[htb!]
  \centering   
  \includegraphics[width=0.7\textwidth]{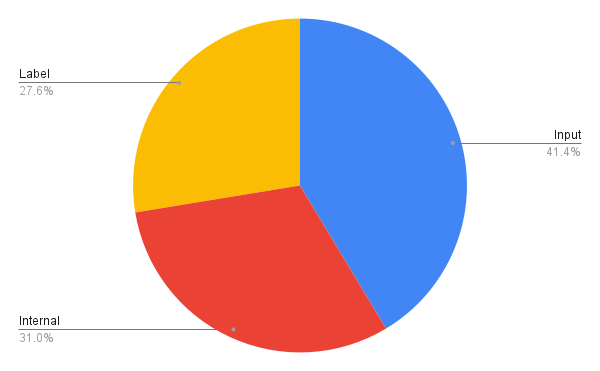}
  \caption{Percentage of the regularization works surveyed in the manuscript.}  
  \label{f.chart}  
\end{figure}

Even though most of the works are applied to the input, there are many studies dedicated to internal structures and the label layer. Figure~\ref{f.chart} depicts the proportion of the scientific works presented in this survey.

Around $44\%$ of the works relies on changes on the input, most known as data augmentation strategies. The easiness of changing parameters and structures in a CNN's input may explain such an amount of works. Image processing- and computer vision-driven applications still play a significant role when dealing with deep learning. The second most common regularization approaches stand for the ones that perform changes in the internal structures. Dropout~\cite{dropout} contributed considerably to advance in this research area. Several works~\cite{maxdropout,localdrop,autodropout} are mainly based on Dropout, while some of them~\cite{shakedrop,shakeshake} are new approaches.

\subsection{Lack of Label Regularizers}
\label{ss.lack_label_regularizers}

We want to highlight the importance of more research on regularizers that work on a neural network label level. Although around $22$\% of the works make changes on the label as a regularization strategy, we found two relevant works on the area only~\cite{tsla,inceptionv2v3}. Some hypotheses may be raised here.

The first one is that the label level is not intuitively changed as the input or in the middle-level of a neural network. Performing changes in both levels is more natural, for it is visually more obvious to understand what is going on during training and inference. However, it is harder to explain what happens when label changes are performed. Even though the original work~\cite{inceptionv2v3} argues that it prevents the overconfidence problem, it fails to explain why such a situation is avoided.

Another explanation is the lack of mathematical explanation for most approaches. Fortunately, some techniques such as Dropout~\cite{cavazza2018dropout} and Mixup~\cite{carratino2020mixup} present interesting insights about their inner mechanism. An algebraic proof that label smoothing works well may be an essential step for the development of new strategies concerning the last level's regularization.

Finally, it is always good to remember that one of the most critical steps for developing a machine learning area is creating reliable-labeled datasets. Although we focused on regularization strategies, it is worth remembering that, eventually, a breakthrough on the way we work with labels may lead to more powerful systems. Therefore, we emphasize that more works related to the label-level regularization are worth researching.

\section{Convolutional Neural Networks}
\label{s.cnn}

Neural networks have been used since the 1950s when the first neuron emulation, called Perceptron~\cite{perceptron}, was developed. However, it can primarily address linearly separable feature spaces. However, in the 1980s, the development of the backpropagation algorithm~\cite{backpropagation} to set new values in a structure that uses several Perceptrons in more than one layer, called the Multilayer Perceptron (MLP), made it possible to solve nonlinear problems as well. Even with these advances, it still lacks some relevant results to solve unstructured data problems, such as images.

\begin{figure}[htb!]
  \centering   
  \includegraphics[width=0.8\textwidth]{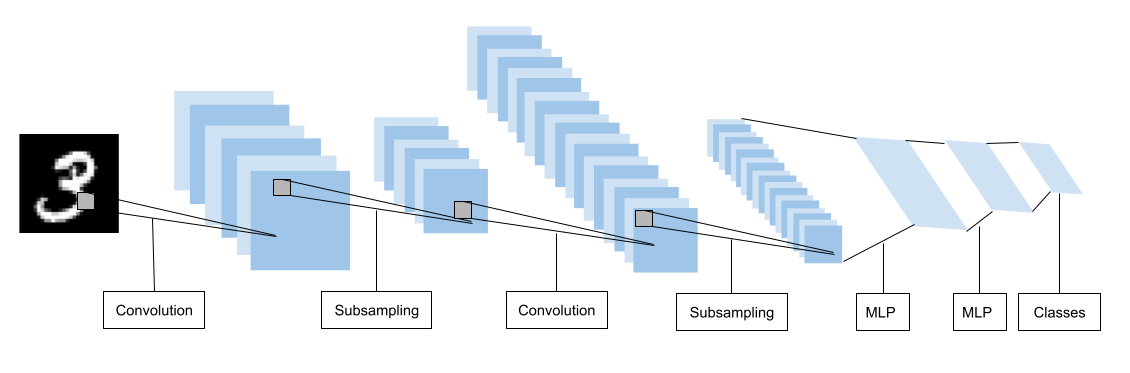}
  \caption{The LeNet-5 structure. Inspired in the picture from~\cite{lenet}.}  
  \label{fig:lenet}  
\end{figure}

In late 1990, a new neuron structure emerged based on the 2D convolution process, the so-called Convolutional Neural Network~\cite{lenet}. The 2D convolution process can find different features on an image, depending on the convolutional kernel's size and values. What makes a CNN so valuable for image processing is the possibility of stacking convolutional processes to find out different features whose training can be accomplished using the well-known backpropagation algorithm. Figure~\ref{fig:lenet} illustrates a standard structure of a Convolutional Neural Network.

Even being so powerful, it still needs lots of data to achieve relevant results, thus requiring considerable computational power. In the middle of the 2000s, GPUs' use accelerated the training process hugely, becoming possible to solve image processing problems in a feasible time. From 2010, the first relevant result emerged. The AlexNet structure~\cite{alexnet} achieved first place in the Image Net classification challenge, overcoming the runner-up result by more than 10\%. It is an $8$-layer CNN with an MLP on top to perform the classification. Since then, other CNN structures have appeared, each one with new features in their structure.

The Visual Geometry Group developed the VGG architecture~\cite{vgg} which demonstrates, for the first time, that stacking convolution layers with smaller kernels perform better than shallow layers with bigger kernels, even then performing over the same region. This architecture achieved first place in the ImageNet classification challenge in 2012. Another architecture family with relevant results is the Inception~\cite{inceptionv1,inceptionv4,inceptionv2v3}, which was developed by Google by parallelizing kernel operations in the same layer and then fusing them before the next layer.

About the same time the first Inception architecture showed up, Microsoft presented the Residual Network, most known as ResNet~\cite{resnet}. It works by fusing the output of layers with the same dimensions before the pooling operation. It looks like a simple operation at first, but later on, it has been shown that this residual connection helps the backpropagation algorithm to handle better the well-known vanishing/exploding gradient shortcoming~\cite{identitymapping}.

Neural Architecture Search, known as NAS~\cite{nasnet}, developed a new way to find better CNN architectures. Using an agent trained by the Q-Learning technique~\cite{qlearning}, it can find out the CNN that can achieve the best result according to some rules. The drawback of this technique is that it takes a considerable amount of time to discover the best neural network architecture. However, recent studies~\cite{fastnas} showed how to improve the search algorithm, making it faster to discover new architectures.

Later in 2018, Google showed the NAS could be improved when some rules are better designed, such as the computational limit, input size, and other parameters, and incorporate other architectures, such as Squeeze-and-Excitation~\cite{squeeze}, ending up in the EfficientNet family~\cite{efficientnet}. The original work showed eight different architectures (called B0-7), which perform using the same quantity of floating points operation (FLOP) as other architectures but achieving better results. In the same study, the EfficientNet architectures delivered state-of-the-art results in five different datasets.

All works discussed until now operate on the image classification problem. However, CNN's can be used in several other tasks. One interesting problem is object detection in natural scenes. The R-CNN~\cite{r-cnn}, for instance, works in two stages, being the first to find interest regions on the image, and the final stage classifies each region in the desired objects. The You Only Look Once, known as YOLO~\cite{yolo}, goes one step further and performs the localization and classification steps in the same stage.

Another task well solved by CNN concerns image reconstruction. In this case, most of them are Fully Convolutional Networks (FCN), which means that every single layer on the neural network is a convolutional layer. One relevant work in this area is the Residual Dense Network, which has a version for super-resolution~\cite{rdn-sr} and image denoising~\cite{rdn-reconstruction} purposes. Another significant development is the DnCNN~\cite{dncnn}, which not only resolves problems for image denoising, JPEG deblocking, and super-resolution but has a version that can solve the three problems without any information about the input image, performing a blind reconstruction.

The Generative Adversarial Network (GAN) was first developed using MLP~\cite{gan}; however, it is used mainly with convolutional layers to solve diverse problems. One problem tackled by GAN is the style transfer, in which the StackGAN~\cite{stackgan} shows a very nice result, being able to change the style completely without losing relevant information. Another work with good results is the ERSGAN~\cite{esrgan}, which deals with the super-resolution of images. The neural network shows outstanding results by training a Residual-in-Residual Dense Network (RRDN) using the GAN approach.

\section{Regularization based on data augmentation}
\label{s.dataaug}

When thinking about changes in the training data for CNNs, maybe the most intuitive way is to perform alterations on the input for all changes can be visualized (or at least imagined) before the beginning of the model's training. Since the first CNN model~\citep{lenet}, basic data augmentation, such as flipping and noise adding,  has proven it can help the trained model generalize better.

\subsection{Cutout}

\begin{figure}[htb!]
  \centering   
  \includegraphics[width=0.8\textwidth]{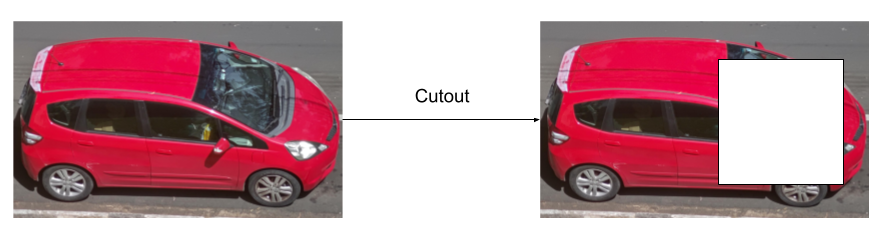}
  \caption{How Cutout works. Extracted from~\cite{cutout}}  
  \label{f.cutout}  
\end{figure}

One straightforward but powerful technique to perform data augmentation is the well-known Cutout~\cite{cutout}. During training, it randomly removes regions of the image before feeding the neural network. In~\cite{cutout}, the authors exhaustly analyzed what would be the ideal size of the removed region in the CIFAR-10 and CIFAR-100 datasets. The ideal size varies according to the number of instances per class and the number of classes for a given dataset. For example, the best results on the CIFAR-10 dataset were accomplished by removing a patch of size $16\times16$, while for CIFAR-100  the region size concerning best results was $8\times8$. For the SVHN dataset, the best crop size was found out by using a grid search, which outputs the $20\times20$ size as ideal. Regarding the STL-10 dataset~\cite{stl} the cut size for the best result was $32\times32$. Figure~\ref{f.cutout} shows how Cutout works.

\subsection{RandomErasing}

\begin{figure}[htb!]
  \centering   
  \includegraphics[width=0.8\textwidth]{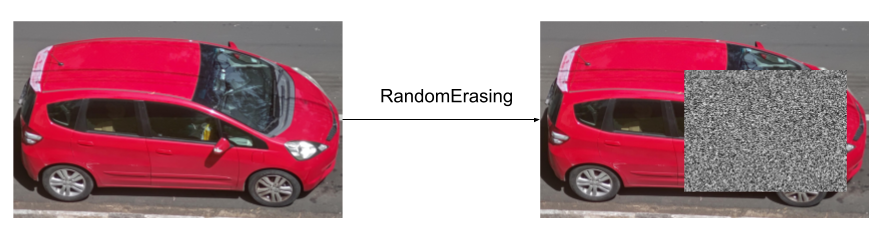}
  \caption{How RandomErasing works. Extracted from~\cite{randomerasing}.}  
  \label{f.random_erasing}  
\end{figure}

RandomErasing~\cite{randomerasing} was further developed based on the Cutout technique. While the latter removes random crops of the image, RandomErasing is concerned about removing and randomly adding information on the blank space, such as noise. Different from Cutout, RadomErasing does not remove pieces of the image every time. In this work, the authors evaluated the method on three different classification datasets (CIFAR-10, CIFAR-100, and Fashion-MNIST), the PASCAL VOC 2007~\cite{pascal-voc-2012} dataset for object detection, and three different CNN architectures for person re-identification (IDE~\cite{ide}, TriNet~\cite{trinet}, and SVDNet~\citep{svdnet}). For the classification task, four different architectures were used for evaluation purposes: ResNet~\cite{resnet}, ResNet with pre-activation~\citep{identity}, Wide Residual Networks~\cite{wrn} and ResNeXt~\cite{resnext}, including four distinct setups for the two first architectures. In all cases, the RandomErasing approach accomplishes a relevant error reduction (at least $0.3$\%). For the object detection task, the mean average precision (mAP) was increased by $0.5$ when the model was trained only with the available data from the dataset and $0.4$ gain when the training data was combined with the PASCAL VOC 2012 training dataset~\cite{pascal-voc-2012}. Figure~\ref{f.random_erasing} shows how RandomErasing works.

\subsection{AutoAugment}

AutoAugment~\citep{autoaugment} tries to find out what transformations over a given data set would increase the accuracy of a model. It creates a search space for a given policy using five different transformations ruled by two additional parameters: the probability of applying a given alteration (which are: Cutout, SamplePairing, Shear X/Y, Translate X/Y, Rotate, AutoContrast, Invert, Equalize, Solarize, Posterize, Contrast, Color, Brightness, and Sharpness) and the magnitude of this change. These policies are then fed into a "child" model, which is a CNN trained with part of the training data set. The accuracy of this CNN is informed to a "controller" model, which is a Recurrent Neural Network (RNN) - more specifically, a Long-Short Term Memory. This RNN outputs the probabilities of a given policy to be used in the future. At the end of the controller training procedure, the five best policies (each one with five sub-policies) are used to train the final model used to evaluate the data set. Using these generated policies and sub-policies, AutoAugment accomplished state-of-the-art results on CIFAR-10, CIFAR-100, SVHN, and ImageNet datasets. One huge advantage of this approach is the transferability of these policies across different datasets: in the original work, the policies found out for ImageNet were used to train five other different datasets, improving the results significantly even when the AutoAugment technique was not trained on them. One disadvantage of this approach is the time used to train the controller model: for the ImageNet dataset, for instance, it took around $15,000$ hours of processing, which may be impracticable in several cases. Fast AutoAugment~\cite{fastautoaugment} aimed at overcoming such a bottleneck with a new algorithm, reducing the time related to the search procedure significantly, besides producing similar results.

\subsection{PBA}

Population Based Augmentation (PBA)~\cite{pba} not only showed a novel augmentation algorithm but demonstrated schedule policies instead of fixed policies, which improves the results from the previous studies~\cite{autoaugment,fastautoaugment}. At every $3$ steps, it changes half the policies, being $1/4$ changes in the weights and the other $1/4$ a change in the hyperparameter. While AutoAugment implies an overhead of $5,000$ hours for training over the CIFAR-10 data set, PBA increases it by only $5$ hours.

\subsection{RandAugment}

As mentioned before, a huge bottleneck for the methods which look for finding the best data augmentation involves their computational burden since it may take longer than the own neural network training. Another problem is related to the strategies found during the search, which may end up in a sub-optimal strategy, i.e., it does improve the results locally; however, it does not lead to the best global result for it uses a shallower neural network and assumes that this rule can be applied to any other, and possibly, deeper architecture. RandAugment~\cite{randaugment} uses the $14$ most common policies found on previous works~\cite{fastautoaugment,autoaugment,pba} and performs the search of the magnitude of each policy during training, thus removing the need for a preliminary exploration step and tailoring the data amplification to the current training CNN. Results show that the method is not only faster than previous approaches~\cite{fastautoaugment,autoaugment,pba} but improves the outcomes significantly.

\subsection{Mixup}

\begin{figure}[htb!]
  \centering   
  \includegraphics[width=0.8\textwidth]{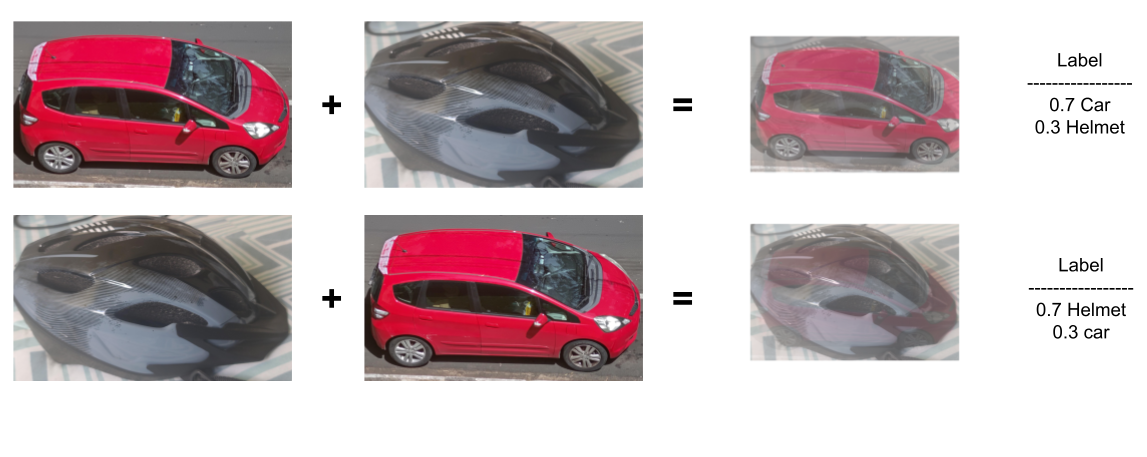}
  \caption{Two different examples using Mixup. Extracted from~\cite{mixup}}  
  \label{f.mixup}  
\end{figure}

One possibility for training CNN concerns mixing two images from the training data set and forcing the model to determine which class this mixture belongs reliably. However, it is not widespread how to generate the encoding label for such a mixture. Providing this new input/output training pair allows the model to learn more features from corrupted inputs. The original work shows that models using such an approach can improve results not only in the image classification task but in speech recognition, stabilization in generative adversarial networks, tabular datasets, and other problems. Figure~\ref{f.mixup} demonstrates how mixup works.

\subsection{CutMix}

\begin{figure}[htb!]
  \centering   
  \includegraphics[width=0.8\textwidth]{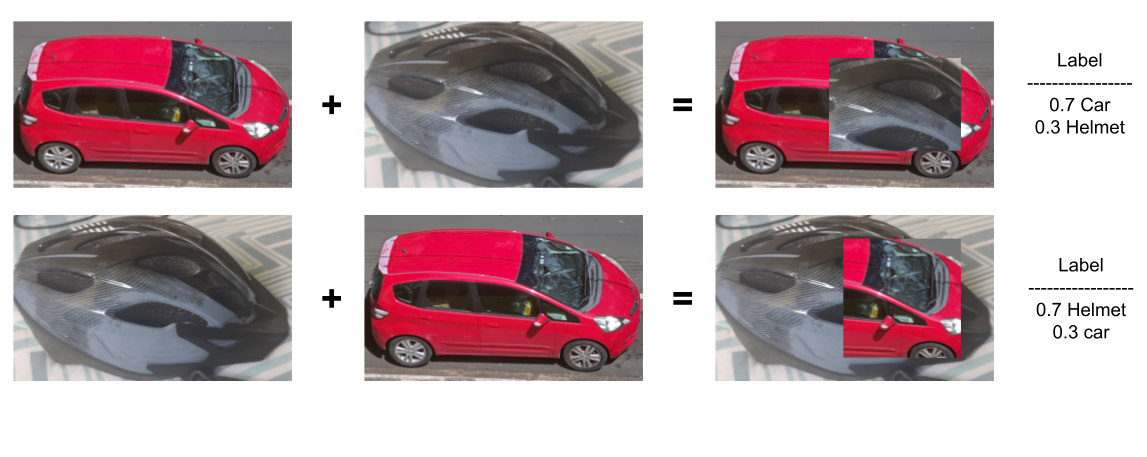}
  \caption{How CutMix works. Extracted from~\cite{cutmix}.}  
  \label{f.cutmix}  
\end{figure}

Another strategy to mix inputs and labels to improve the results is the CutMix~\cite{cutmix}. Unlike Mixup, CutMix replaces entire regions from a given input and changes the label by giving the same weights as the area used by each class. For example, if a cat's image is replaced in $30$\% by an image of an airplane, the label is set to be $70$\% cat and $30$\% airplane. This strategy shows a significant improvement in results. By using techniques that map the most activated regions (e.g., grad-CAM~\cite{gradcam}), one can observe that the generated heat maps highlight better the areas that define the object of interest more accurately. Figure~\ref{f.cutmix} illustrates the technique.

\subsection{CutBlur}

Several Deep Learning tasks targeting image processing, such as image classification or object detection, can improve their models by using data augmentation. Several works, such as AutoAugment~\cite{autoaugment,fastautoaugment}, Cutout~\cite{cutout}, and RandomErasing~\cite{randomerasing} can improve results significantly by applying some clever but straightforward transformations on the training images. However, for super-resolution (SR) tasks, the literature lacks works that proposed regularization techniques to handle the problem explicitly. Even though the aforementioned techniques can be used and possibly improve results, they are not natively designed to cope with the SR problem. The only approach found, so far, is the CutBlur~\cite{cutblur}, which works by replacing a given area on the high-resolution image (HR) with a low resolution (LR) version from a similar region. The authors showed that CutBlur helps the model generalize better on the SR problem but that the same technique can be applied to reconstruct images degraded by gaussian noise.

\subsection{BatchAugment}

One important hyperparameter for training CNNs concerns the mini-batch size, which is used to calculate the gradient employed in the backpropagation. Usually, the GPU's upper limit is employed for such a hyperparameter, which is crucial to speed up the convergence during training. The Batch Augmentation work~\cite{batchaugment} cleverly uses this limit. Instead of just fulfilling the entire memory with different instances from the dataset, it considers half of the memory limit using the default set up for data augmentation and then duplicates all instances with different data augmentation possibilities. It sounds like a straightforward technique; however, results demonstrate that neural networks that use such an approach have a significant improvement on the final results. Another point is that, by duplicating the augmented images, the analysis showed that it is necessary fewer epochs for convergence.

\subsection{FixRes}

The image resolution may influence both the training step efficiency and the final classification accuracy. For instance, the research on EfficientNet~\cite{efficientnet} highlights this idea by making the input size one of the parameters that influence the final result. However, if a model is trained, for example, with a resolution of $224\times224$, the test set's inference uses the exact resolution. The work proposed by~\cite{fixres} highlighted that the resolution of the test set should be higher than the resolution used for training. This change not only produces a more reliable neural network but it trains faster than the traditional approach, for it requires less computational effort for training since the images used for such a purpose are smaller than the ones used for inference. The proposed approach shows it can improve the results on other datasets when transfer learning is used.

\subsection{Bag-of-Tricks}

One critical point of the works analyzed here is that they frequently do not combine any other regularizer with their current research. Hence, it is hard to know how two regularizers can influence one another. The Bag of Tricks research~\cite{bag} performs this investigation by combining several known regularization methods, such as Mixup~\cite{mixup}, Label Smoothing~\cite{inceptionv2v3} and Knowledge Destilation~\cite{distill}. The ablation study shows that if some cleverness is applied, the final result can be significantly improved. For instance, a MobileNet~\cite{mobilenets} using this combination of methods improved its results by almost $1.5$\% in the ImageNet dataset, which is a significant gain. However, the research lacks a deeper evaluation of methods for regularization among layers, such as Dropout~\cite{dropout}.

\section{Regularization based on internal structure changes}
\label{s.internal}

Regularization methods can work in different ways. In this paper, we define internal regularizers as the ones that change the weights or kernel values during training without any explicit change on the input. This section is divided into two main parts: the first presents a deeper description of how dropout works and some of its variants, such as SpatialDropout and DropBlock. In the second part, we describe other methods that aim to perform other tensors' operations, such as Shake-shake regularization.

\subsection{Dropout and variants}
\label{ss.dropout_variants}

Dropout~\cite{dropout} was proposed as a simple but powerful regularizer that aims to remove some neurons, thus forcing the entire system to learn more features. The original work shows it can be applied not only on CNNs but in Multilayer Perceptrons (MLPs) and Restricted Boltzman Machines (RBMs). The probability of dropping out each neuron is estimated through Bernoulli's distribution at each step of the training phase, thus adding some randomness in the process. The original work shows dropped neural networks can generalize better than standard ones.

\subsection{MaxDropout}

While Dropout~\cite{dropout} randomly removes the neurons in the training phase, Maxdropout~\cite{maxdropout} deactivates the neurons based on their activations. It first normalizes the  tensor's values and then sets to $0$ every single output greater than a given threshold $p$, so the higher this value, the most likely it to be deactivated. The original work shows it can improve ResNet18 results on CIFAR-10 and CIFAR-100~\cite{cifar} datasets, and it also outperforms Dropout on the WideResNet-28-10 model~\cite{wrn}.

\subsection{DropBlock}

CNN works so well on images and related fields (such as video) that it can generate correlated regions among its neurons. For instance, in the image classification task, heatmaps generated by techniques like grad-CAM~\cite{gradcam} show that, when correctly estimated, CNN highlights regions of interest around the object that has been classified. DropBlock~\cite{dropblock} shows that removing entire areas of a given tensor (i.e., feature map) can help the model to generalize better. By using ResNet-50 and AmoebaNet-B models on the image classification task, RetinaNet on object detection, and ResNet-101 for image segmentation,  it shows that it can improve results better than Dropout and other internal regularizers~\cite{autoaugment,spatialdropout}. DropBlock is applied on every feature map of the CNN, starting the training with a small ratio and slowly increasing its value. Its experiments show relevant results on the ImageNet dataset, increasing the baseline accuracy by almost $2$\% when using ResNet-50, beating other regularizers, such as Cutout and AutoAugment, and around $0.3$\% when using AmoebaNet-B. In the object detection task, the RetinaNet model is improved by more than $1.5$ in the AP metric. 

\subsection{TargetDrop}

Attention mechanism can be incorporated into a given regularizer so it can act in the appropriate region. For instance, the TargetDrop~\cite{targetdrop}  combines this mechanism with DropBlock. During training, it allows the entire system to remove most discriminative areas on a given channel. Results show this method not only accomplishes better results than DropBlock but, by using grad-CAM~\cite{gradcam}, demonstrates more consistency in the region that determines to which class a given input belongs.

\subsection{AutoDrop}

Although effective, Dropout lacks spatial information for choosing what neuron to drop. DropBlock's strategy is to drop entire random regions on hidden layers instead of singular neurons, thus forcing a CNN to learn better spatial information. However, the drop pattern is manually designed and fixed, which may be improved if these patterns could be learned during training. AutoDrop~\cite{autodropout} forces the CNN to learn the best drop design according to information from training by using a controller that learns, layer by layer, the best drop pattern. Results in CIFAR-10 and ImageNet show that these patterns improve results and can be transferred in between data sets.

\subsection{LocalDrop}

The Rademacher complexity was used to redefine both Dropout and DropBlock~\cite{localdrop}. After an extensive mathematical analysis of the problem, a new two-stage regularization algorithm was proposed. Although very time-consuming, the proposed method achieves relevant improvement on different CNN architectures targeting image classification.

\subsection{Other methods}
\label{ss.other_methods}

In the last few years, the use of residual connections, first introduced in the well-known neural architecture ResNet~\cite{resnet}, and their further improvements~\citep{resnext,wrn} have achieved relevant results on several tasks. Later studies~\citep{identity} have shown that such a success is due to the creation of a structure called ``identity mapping", which is the reconstruction of the original input. The residual connection then forces the model to learn how to construct these structures. 

\subsection{Shake-Shake}

One way to force regularization on these architectures is to give different weights to each branch of the residual connections during training. The original ResNets works by adding the weights on each branch without any differentiation. During training, Shake-shake~\citep{shakeshake} works on 3-branch ResNets by changing the multiplication factor of each branch on the forward pass and multiplying by a different value on the backward pass, thus changing how each branch affects the final result. For the inference, it multiplies each branch by a factor of $0.5$. 

Results on CIFAR-10 show that such an approach can improve outcomes by at least $0.15$\%, achieving almost $0.6$\% improvement on the best result. Results on the CIFAR-100 were improved by $0.4$\%; however, in this specific case, the removal of weights' changes on the backward pass ends up in slightly better results, improving by $0.5$\%. Besides the improvement, this method only works on 3-branches ResNet, making it hard to compare other methods directly.

\subsection{ShakeDrop}

One improvement to tackle the problems of Shake-shake is the ShakeDrop~\citep{shakedrop}. It works not only on ResNeXt architecture but on ResNet, Wide ResNet, and PyramidNet too. To accomplish such results, ShakeDrop changes the formulation proposed by Shake-shake. The combination of these perturbations on the branches shows ShakeDrop has more tools not to be trapped on local minima. Results show that it can outperform the original results obtained by each architecture mentioned earlier. 

\subsection{Manifold Mixup}

A neural network is usually generalized as a function that, given input data and a set of learnable parameters, outputs the target value accordingly. The Manifold Mixup~\cite{manifoldmixup} acts like the Mixup~\cite{mixup}, however, operating in any internal layer of a CNN, and not only in the input layer. A deep neural network can be considered a set of smaller neural networks. Each one outputs some desired features; therefore, if all sub-nets work well, the final result can be regarded as a good one. Yang et al.~\cite{gradaug} propose a new strategy to design the loss function: it first calculates the traditional loss of a mini-batch through the feedforward process. After that, it generates sub-networks from the original one and then computes one loss for each model by supplying the same mini-batch using different image transformations. Finally, the final loss is calculated by adding the traditional loss with the losses from each sub-network. This technique shows a great potential improvement in different datasets and CNN architectures.

\section{Label Regularization}
\label{s.label_smoothing}

Revisiting some information on Table~\ref{tbl:methods}, other methods use label smoothing as part of their regularization strategy. For instance, Mixup~\cite{mixup} averages the values of the labels depending on the interpolation between two different images. The same rule is applied for the Manifold Mixup technique~\cite{manifoldmixup}; however, the data interpolation is computed among the layers and the same calculus is used for resetting the label values.

Another regularizer that uses label transformation is Cutblur~\cite{cutblur}. In this case, the transformation is used so wisely that, during training, the label could be inverted with the input, making the input as the label, and the model would converge as expectedly. The reason for this expected result is due to the cut size of the low-resolution and high-resolution images, which are not defined beforehand. It means that the input can be a low-resolution image with a crop from the high-resolution image, and the label would be the high-resolution image with the crop from its low-resolution counterpart. Therefore, inverting the label and input still makes sense.

Other methods can also have their results improved by using some rationale borrowed from label smoothing. For instance, Cutout~\cite{cutout} removes parts from the input, so it makes sense to "remove" part of the label according to the crop size as well. Pretend the crop size is $25$\% of the image, so the active class could be dropped from $1$ to $0.75$. The same strategy can be applied to RandomErasing~\cite{randomerasing}. Methods that drop neurons during training, such as Dropout~\cite{dropout} could, for example, drop the values of the hot label by the same range of the total active neurons deactivated during training.

\subsection{Label Smoothing}

It is widespread in a general classification task to use the one-hot vector to encode the labels. Dating back from $2015$~\cite{inceptionv1}, label smoothing proposes a regularization technique in the label encoding process by changing the value on each position of the hone-hot representation.

Label smoothing works by preventing two main problems. First, the well-known overfitting, i.e., the situation where the model learns the information about the training set but cannot generalize the classification in the test set. The second and less obvious is overconfidence. According to the authors~\cite{inceptionv1}, by using the smoothing factor over the encoding label, the softmax function applied over the vector produces values closer to the smoothed encoded vector, limiting the value used in the backpropagation algorithm and producing a more realistic value according to the class.

\subsection{TSLA}

One difficulty of using label smoothing is to find out what value of $\epsilon$ (i.e., smoothing factor) is the ideal, either for a general or for a specific data set. The original work suggests that $\epsilon = 0.1$ is the excellent condition; however, the Two-Stage Label Smoothing (TSLA)~\cite{tsla} suggests that, in general, the gradient descent combined with the label smoothing technique can only improve the results until a certain point of training, after that it is better to set all values to $0$ and $1$ for the active class. For instance, when training the ResNet18 in the CIFAR-100 data set for $200$ epochs, results suggest the best performance is achieved when label smoothing is used until the epoch $160$.

\subsection{SLS}

\review{Usually, it is not straightforward to define appropriate values for the label smoothness factor. Structural Label Smoothing (SLS)~\cite{sls} proposes to compute such a value by estimating the Bayes Estimation Error, which, according to authors, helps define the label's boundaries for each instance. Several experiments show that this approach can overcome the traditional label smoothing method on different occasions. Although the work is fully evaluated on MobileNet V2~\cite{mobilenetv2}, it does not consider other neural network architectures. Even though some popular data sets were used for comparison purposes, e.g., CIFAR and SVHN, the work is limited to MobileNet-V2 only.}

\subsection {JoCor}

\review{This work proposes a new approach to avoid the influence of noisy labels on a neural networks. JoCoR~\cite{jocor} trains two similar neural networks on the same data set and tries to correlate two different labels. The method calculates the loss by adding the cross-entropy losses of both networks plus the contrastive loss between them and then uses only the most negligible losses on the batch to update the parameter of the architectures. The authors argue that both networks agree with the predictions by using the smallest values to update parameters, and the labels tend to be less noisy. Although the method was developed for weakly supervised problems, it could easily fit traditional supervised problems, such as data classification, to improve outcomes. The downside of this method is using two neural networks for training, which requires more processing and memory.}

\section{Methodology}
\label{s.methodology}

We provide a direct comparison among each regularizer described in this work. We divided each table by model for a more transparent comparison and then provided the result for each dataset available on the original work and related works. The results are shown on the classification task, showing how each algorithm performed in the most common datasets.

\subsection{Datasets}
\label{ss.datasets}

The last important part of training a neural network and defining the baseline is to decide what dataset should be used for evaluation, either for training and validation. For the sake of research in regularization in image processing, two datasets are considered the most frequent, i.e., CIFAR, Imagenet\review{, and SVHN}.

\subsubsection{CIFAR}
\label{ss.cifar}

The original CIFAR (Canadian Institute For Advanced Research) dataset consisted of $80$ million images; however, due to some ethical problems, such as offensive and prejudicial images, the authors decided to make it unavailable~\footnote{More information: http://groups.csail.mit.edu/vision/TinyImages/}. Instead, two other subsets have been frequently used in regularization research: (i) CIFAR-10 and (ii) CIFAR-100.

The CIFAR-10 subset is compounded by $60,000$ $32\times32$ images, divided into $50,000$ figures for training and the remaining $10,000$ for test/validation. It is divided into ten classes between animals (bird, cat, deer, dog, frog, and horse) and objects (airplane, automobile, ship, and truck). 

The CIFAR-100 is also built by $60,000$ $32\times32$ images, divided into $50,000$ figures for training and the remaining $10,000$ for test/validation. However, it is divided into $100$ classes, being harder to classify than its counterpart version. Objects and animals also compound the classes. Besides, both versions of the CIFAR dataset use the same images for training and testing.

\subsubsection{ImageNet}
\label{ss.imagenet}

Ordinarily called a ``dataset", the ImageNet is a project developed to improve artificial intelligence tasks, such as image classification. ImageNet dataset is usually associated with the ``2012 ImageNet Large Scale Visual Recognition Challenge" (ILSVRC), which comprises $1,240,000$ $224x224$ images for training and $50,000$ for validation purposes divided into $1,000$ classes. It has one order of magnitude bigger than the CIFAR subsets.

ImageNet is historically relevant in the deep learning community, mainly for those who work with image classification, for it was in the 2010 ILSRVC that the first practical work using CNN was demonstrated: the AlexNet~\cite{alexnet}, a neural network compounded by convolutional and a multilayer perceptron (MLP) layer, trained end-to-end, achieved first place in the context, outperforming the runner-up by more than 10\% in the accuracy metric. Since then, several deep learning architectures have been designed to cope with image classification problems on that dataset~\cite{vgg,inceptionv1,resnet,efficientnet}.

\subsubsection{SVHN}
\label{ss.svhn}

Used less frequently than the datasets mentioned above, but it is still a good baseline, the Street View House Numbers (SVHN)~\cite{svhn} is a set of images that comprises houses' numbers. Some characteristics come from the MNIST data set~\cite{mnist}, like the 0-9 digits as the label; however, it has another order of magnitude of difficulty and number of instances.

According to the website~\footnote{http://ufldl.stanford.edu/housenumbers}, there are two versions of the dataset. The first one is a collection of images containing two or more digits in the same number, forming more complex instances and not being frequently used. All images are colored and vary in resolution and size.

Regularization works often uses the second version of the dataset. The same images from the first set are once more used; however, they are now segmented by each digit, so every label is among the 0-9 range. These images are scaled in $32\times32$ size, varying in resolution and color. There are three divisions of the dataset. The first is the original training set, formed by $73,257$ labeled instances. The second division contains $26,032$ images, and it is used for evaluation purposes. The third subset is called ``extra" and includes $531,131$ designated figures. Realize that some works use the ``extra" and ``training" parts for training models, and others use the ``training" portion to the size and time taken for training. Results reported in Table~\ref{t.classification} are the ones that use both sets in training.

\subsection{Architectures}
\label{ss.architectures}

For a fair comparison, two regularization methods must use the same architecture. In all works mentioned earlier, at least one of the architectures described in this subsection is used.

\subsubsection{ResNet}
\label{sss.resnet}

The oldest architecture used in most of the regularization works, the ResNet family~\cite{resnet} is still one of the most commonly used CNNs. It stands for the first neural network to use residual connections, which is the concatenation of the output from previous layers with further transformations. The residual connection is powerful, functional, and easy to implement.

Several works use two variants of the ResNet. These variants are different not only because of the depth of the neural network, but the blocks have a different constitution. The ResNet-18, as the name suggests, is built from $18$ layers with residual connections between every block, each block having two or three layers of a sequence of convolution and batch normalization~\cite{batchnorm}, depending on the position of the block, with the third layer as a pooling layer by changing the stride of the convolution to $2$ instead of $1$. The other variant is the ResNet-50, which uses $50$ layers; however, built-in more complex blocks, the so-called ``bottleneck". Every block has three or four layers of convolution and batch normalization, again depending on the block's position. The fourth block works as a pooling layer, with the same rules as previously described.

\subsubsection{Wide Residual Network - WRN}
\label{sss.wrn}

Another widespread architecture in regularization works is the Wide Residual Network (WRN)~\cite{wrn}. It uses the same concept of residual connection between layers, but it has some structural differences from ResNet. The first one is the use of the concept of pre-activation (Pre-Act) layers. Both ResNet-18 and ResNet-50 use a sequence of convolution, batch normalization, and ReLU activation in their blocks. The Pre-Act block changes this sequence, i.e., it first employs batch normalization, then ReLu activation, and finally the convolution over the input. As shown in the original work~\cite{identitymapping}, this sequence can outperform the traditional chain.

The second difference is the change in the widening and depth of the neural networks. It is widespread to observe the WRN being called ``WRN-\textit{d-k}", with $k$ as the widening factor and $d$ is the depth factor. The depth is the usual concept, i.e., it means the number of convolutional layers; however, the widening changes the structure significantly. When $k = 1$, it has the same structure as the ResNet; however, it means the layer has more convolutional kernels in a given layer when this number increases. This small change can generate a much shallower network (with $16$ layers) with similar results as the ResNet-1001, containing $1,001$ layers. In the regularization works, the most common architecture is to employ the WRN-28-10, but it is possible to find some of them using the WRN-16-8 either.

The last distinction is the use of Dropout in the original architecture. The number of convolutional layers increases drastically on each layer, which may lead to overfitting~\cite{wrn}. Dropout regularization between the convolutional layers after the ReLU activation helps perturb the batch normalization operation, which prevents overfitting.

\subsubsection{ResNeXT}
\label{sss.resnetx}

Intuitively, increasing the number of layers, blocks, or the number of kernels on some layers leads to the idea of better final results. For instance, some studies increase the number of blocks~\cite{identitymapping} or the number of convolution kernels in the layers~\cite{wrn} heavily. ResNeXt~\cite{resnext} introduces the concept of cardinality to accomplish better results.

Since ResNet~\cite{resnet}, most of the neural networks are composed of the main branch, i.e., convolutional, activation, and batch normalization operations, followed by residual connections. In the ResNeXT architecture, the main string is divided by its cardinality value: for example, if a ResNet has a branch with $32$ convolutions, followed by $64$ and then another $32$ convolutions, a ResNeXT block with cardinality $32$ divides the main branch in $32$ streams of $1$, $2$, and $1$ convolution processes, and then concatenate all units before adding the residual value. It looks just a tiny difference in the general design; however, results in CIFAR and ImageNet datasets show that such a remodeling leads to better results. Comparing to previous architechtures~\cite{wrn,resnet,identitymapping}, it showed better outcomes.

\subsubsection{PyramidNet}
\label{sss.pyramidnet}

The last most common neural network is the one that achieves the best general results among the four mentioned here. The PyramidNet~\cite{randaugment} shows some new procedures to improve outcomes from previous convolutional neural networks. The first difference is the size of each residual block. While most neural networks either keep the size of the output or downsample it and increase the feature map in the following layer, the PyradmidNet gradually increases the dimensionality in the subsequent layer. Such a procedure has been shown to improve the results in the classification task.

Such an increase in the feature map's size can also occur inside a residual block. It means that adding outcomes from the residual branch can be a problem concerning the dimensionality of the input tensor. To solve this, the authors proposed a Zero-Padded Shortcut Connection, which adds the values from a previous smaller tensor into a bigger one. According to He et al.~\cite{identitymapping}, this operation might influence the gradient value because any change in this branch (even a scalar multiplication or a dropout regularization) might lead to wrong backpropagation values; however, the study shows that a zero-padded shortcut does not influence the values because no other operation is performed in the residual connection.

The last improvement is a new residual building block. This study shows that better results can be accomplished if the building block uses fewer ReLU activations. The first ReLU activation of the block does not influence that much in the nonlinearity of the system so that it can be removed.

\section{Experimental Results}
\label{s.experiments}

Convolutional Neural Networks are usually designed to achieve the best possible performance in image processing, depending on the targeting difficulty. Sometimes, the same basic structure can be used in two or more problems, i.e., one needs to change the output layer according to the labels. For instance, the EfficientNet structure~\cite{efficientnet} is re-used in the Efficient-Det work~\cite{efficientdet}. Concerning regularization techniques, other components can be tricky to get rid of. Table~\ref{t.classification} shows the results of several models on CIFAR-10, CIFAR-100, SVHN, and ImageNet datasets. The next sections overview an in-depth discussion about the experiments considered in this paper.

\begin{longtable}{lrp{2.2cm}p{2cm}p{2cm}}

\toprule
        & \multicolumn{2}{c}{Classification datasets.}        &                 &       \\
Method                  & CIFAR-10    & CIFAR-100         & SVHN      & ImageNet  \\ \midrule
ResNet18~\cite{cutout}                  & $4.72\pm 0.21$        & $22.46\pm 0.31$       & -           &    -    \\
+ Cutout~\cite{cutout}                & $3.99\pm 0.13$        & $21.96\pm 0.24$       & -           &    -    \\
+ RandomErasing~\cite{randomerasing}        & $4.31 \pm 0.07$     & $24.03 \pm 0.19$      & -           &    -    \\
+ MaxDropout~\cite{maxdropout}        & $4.66 \pm 0.14$     & $21.93 \pm 0.07$      & -           &    -      \\
+ MaxDropout + Cutout~\cite{maxdropout}   & $3.76 \pm 0.08$     & $21.82 \pm 0.13$      & -           &  -        \\ 
 + Mixup~\cite{mixup}         & $4.2$     & $21.1$      & -           &  -        \\ 
 + MM~\cite{manifoldmixup}        & $2.95 \pm 0.04$     & $20.34 \pm 0.52$      & -           &   -       \\ 
 + TSLA~\cite{tsla}     & -     & $21.45 \pm 0.28$      & -           &  -        \\ 
 + TargetDrop~\cite{targetdrop}     & $4.41$    & $21.37$      & -            &  -        \\ 
 + TargetDrop + Cutout~\cite{targetdrop}    & $3.67$    & $21.25$      & -            &    -      \\ 
 + LocalDrop~\cite{localdrop}   & $4.3$   & $22.2$      & -       & $20.2$ \\ \midrule
WRN~\cite{wrn}            & $4.00$              & $19.25$           & -           & $21.9$        \\
 + Dropout~\cite{wrn}         & $3.89$      & $18.85$           & $1.60 \pm  0.05$      & -       \\ 
 + MaxDropout~\cite{maxdropout}     & $3.84 $       & $18.81$         &  -      &   -       \\
 + TargetDrop~\cite{targetdrop}     & $3.68 $       & -         &  -      &    -      \\
 + GradAug~\cite{gradaug}     &       &      $16.02$     &  -       &  -        \\
 + Dropout + Cutout~\cite{cutout}     & $3.08 \pm 0.16$     & $18.41 \pm 0.27$      & $1.30 \pm 0.03$       &   -       \\
 + Dropout + PBA~\cite{pba}     & $2.58 \pm 0.06$     & $16.73 \pm 0.15$      & $1.18 \pm 0.02$       &   -       \\
 + Dropout + RE~\cite{randomerasing}      & $3.08 \pm 0.05$     & $17.73 \pm 0.15$      & -       &   -       \\
 + Dropout + BA + Cutout~\cite{batchaugment}    & $2.85$    & $19.87$      & -          &  -        \\
 + ShakeDrop~\cite{shakedrop}       & $4.37$      & $19.47$           & -           & -         \\ 
 + Dropout + RE~\cite{randomerasing}    & $3.08 \pm 0.05$     & $17.73 \pm 0.15$      & -           &  -        \\
 + Dropout + Mixup~\cite{mixup}   & $2.7$     & $17.5$      & -           &     -     \\ 
 + Dropout + MM~\cite{manifoldmixup}    & $2.55 \pm 0.02$     & $18.04 \pm 0.17$      & -           & -         \\
 + Dropout + Fast AA~\cite{fastautoaugment}   & $2.7$     & $17.3$      & $1.1$           & -         \\
 + Dropout + RA~\cite{randaugment}    & $2.7$     & $16.7$      & $1.0$           &   -       \\
 + AutoDrop~\cite{autodropout}    & $3.1$     &       & -           & -         \\ 
 + AutoDrop + RE~\cite{autodropout}   & $2.1$     & -      & -            &   -       \\ \midrule
 ResNeXt~\cite{resnext}       & $3.58$    & $17.31$      & -            &      $19.1$   \\
 + RE~\cite{randomerasing}      & $3.24 \pm 0.04$     & $18.84 \pm 0.18$      & -           &  -        \\
+ FixRes~\cite{fixres}      & -     & -     & -       & $13.6$ \\ 
 + ShakeDrop~\cite{shakedrop}     & $3.67$      & $17.80$   & -           & $20.34$         \\
 + Shake-Shake~\cite{shakeshake}      & $3.08 \pm 0.05$     & $17.73 \pm 0.15$      & -           &    -      \\
 + Shake-Shake + Fast AA~\cite{fastautoaugment}     & $2.0$     & $14.9$      & -           &   -       \\
 + Shake-Shake + Cutout~\cite{cutout}   & $2.56 \pm 0.07$     & $15.20 \pm 0.21$      & -           &   -       \\
 + Shake-Shake + PBA~\cite{pba}   & $2.03 \pm 0.11$     & $15.31 \pm 0.28$      & $1.13 \pm 0.02$           &   -       \\
 + Shake-Shake + AA~\cite{autoaugment}  & $2.0  \pm 0.1$    & $14.30 \pm 0.2$       & $1.0$           &  -        \\ \midrule
ResNet-50~\cite{dropblock}        & -     & -     & -       & $23.49 \pm 0.07$ \\
 + ShakeDrop~\cite{shakedrop}     & -     & $25.26$   & -       & -     \\
 + Bag of Tricks~\cite{bag}     & -     & -     & -       & $21.67$ \\
 + GradAug~\cite{gradaug}     & -     & -     & -       & $20.33$ \\
 + LocalDrop~\cite{localdrop}   & $5.3$   & $26.2$      & -       & $21.1$ \\
+ Dropout~\cite{dropblock}      & -     & -     & -       & $23.20 \pm 0.04$ \\
+ Cutout~\cite{dropblock}     & -     & -     & -       & $23.48 \pm 0.07$ \\
+ AA~\cite{dropblock}     & -     & -     & -       & $22.4$  \\
+ BA~\cite{batchaugment}      & -     & -     & -       & $23.14$   \\
 + Fast AA~\cite{fastautoaugment}     & -     & -     & -       & $22.37$   \\
+ Mixup~\cite{mixup}      & -     & -     & -       & $22.1$  \\
+ DropBlock~\cite{dropblock}      & -     & -     & -       & $21.65 \pm 0.05$ \\ 
+ FixRes~\cite{fixres}      & -     & -     & -       & $17.5$ \\ 
+ RA~\cite{randaugment}     & -     & -     & -       & $22.4$ \\
+ AutoDrop~\cite{autodropout}     & -     & -     & -       & $21.3$ \\
+ AutoDrop + RA~\cite{autodropout}      & -     & -     & -       & $19.7$ \\\midrule
PyramidNet~\cite{pyramidnet}        & $3.48  \pm 0.20$    & $17.01 \pm 0.39$       & $1.0$            &   -   $19.2$    \\ 
+ GradAug~\cite{gradaug}      &     & $13.76$         & -           & $20.94$       \\ 
 + ShaekDrop~\cite{shakedrop}     & $3.08$    & $14.96$         & -           & $20.94$       \\ 
 + ShaekDrop + Cutout~\cite{autoaugment}    & $2.3$   & $12.2$          & -           & -       \\ 
 + ShaekDrop + AA~\cite{autoaugment}  & $1.5 \pm 0.1$   & $10.7 \pm 0.2$        & -           & -     \\ 
 + ShaekDrop + Fast AA~\cite{fastautoaugment} & $1.8$   & $11.9$        &             & $20.94$       \\
 + ShaekDrop + RA~\cite{randaugment}  & $1.5$   &    -    &     -       & $15.0$      \\
 + ShaekDrop + PBA~\cite{pba} & $1.46 \pm 0.07$   &    $10.94 \pm 0.09$    &    -       & $15.0$      \\

\bottomrule

\caption{Error (in \%) for each classification dataset using different methods and models. The following acronyms were used: MM = ManifoldMixup, PBA = Population Based Augmentation, RE = RandomErasing, BA = BatchAugmentation, AA = AutoAugment, Fast AA = Fast AutoAugment, and RA = RandAugment.}
\label{t.classification}
\end{longtable}

\subsection{State-of-the-art Regularizer?}
\label{ss.state_of_the_art}

Defining the best regularization technique is not something trivial. For example, the baseline for determining the best image classifier is the one that achieves the best results in the 2012 ILRSVC image classification challenge dataset. During this work, the current research with the best impact on the mentioned dataset is the Meta Pseudo Labels approach~\cite{mpl}. One may argue that the best result achieved by a regularization technique on a given architecture might be considered the best regularization method.

According to Table~\ref{t.classification}, which is a compilation of the results in the most common architectures, one can observe that AutoAugment performs better than PBA using ResNeXT architecture plus Shake-shake regularization in the CIFAR-10 dataset. However, when both regularization algorithms are compared using PyramidNet and ShakeDrop regularization, the opposite happens: PBA achieves better results on CIFAR-10 than AutoAugment. Further analysis showed it is possible to observe other variations in the results.

The best possible assumption about state-of-the-art regularization is based on the results and sorting them into work areas. For example, the likely best regularizer concerning the input layer is the RandAugment, for it does not affect the time spent for training and achieves satisfactory results. For internal regularizers, it is even more challenging. Take ShakeDrop as an example. It has not been evaluated within ResNet-18, while MaxDropout was not assessed for the PyramidNet. Based only on a guess, ShakeDrop appears to have the best results in this particular part. Unfortunately, there are only two regularizers that work directly on labels. For this reason, the TSLA might be considered the best one to be used on a label level.

\subsection{Defining a basic protocol}
\label{ss.basic_protocol}

There are several aspects to be considered for a fair evaluation of a new regularizer. The primary purpose of using regularization is to improve a given baseline architecture by using some operations in the input data, among layers, or in the label. However, a slight difference in the training protocol may infer a better result, not necessarily related to the operations from the regularizer. Another protocol can be removing any other regularization method, even small data augmentation and weight decay. As such, it is possible to verify how a new regularizer can improve a baseline architecture without any other influence.

Some papers~\cite{cutout,randomerasing,maxdropout} train ResNet-18 using the same data transformations, i.e., random flipping, padding pixels, and using the same values for the weight decay. Some works use the same neural network, claim to have some relevant results but do not make the source code available, turning the evaluation process not trustful since there might be other transformations working on training, such as Dropout~\cite{dropout}. Therefore, the primary condition to be cited in this survey is to have the source code available so that it can be compared to other methods directly.

On the other hand, it might be crucial for different reasons to use more than one regularizer in the same evaluation. For instance, the Wide Residual Network~\cite{wrn}, a common architecture used for evaluating new regularizers, has in its layers a dropout regularization. Therefore, wherever a new regularization is proposed (in the input, among layers, or in the label), it should be able to work with the dropout regularization. Another point is that some regularizers incorporate other techniques naturally. For instance, the AutoAugment~\cite{autoaugment} and the Fast AutoAugmentat~\cite{fastautoaugment} incorporate as one of their policies the Cutout~\cite{cutout}. Therefore, a new regularization technique should be able to work with another regularizer and improve the results when both are used together.

\subsection{Use of minor architectures}
\label{ss.minor_architectures}

As a general rule, regularization adds little overhead during training time (AutoAugmentation~\cite{autoaugment} is, perhaps, the only one that increases training time to find out the better policies for data augmentation) and no overhead at all at inference. For this reason, the use of regularizers for avoiding early overfitting of a neural network is strongly recommended. Still, it should be encouraged, sometimes, to use more than one at the same time. No matter what the problem is, everyone has the desire to improve results; however, it is particularly necessary for shallower neural networks.

One point missing in all works analyzed in this survey is the lack of proper investigation concerning lightweight CNNs. Architectures like MobileNet-V3~\cite{mobilenets} should be boosted in regularization works for these smaller designs usually have fewer parameters or make use of less complex calculations. In the same direction, quantization~\cite{survey-quantization} should be dissected to know how a given regularization algorithm influences either training a quantized neural network and performing the quantization after training.

EfficientNet~\cite{efficientnet} provides a clever calculation for defining how an efficient CNN architecture should be designed, based on the width, depth, and resolution. However, for faster and less resourceful hardware, this calculation presents better results when neural networks's resolution and depth are designated as more important than width. It is possible to verify that in the TinyNet work~\cite{tinynet}. It might be a good idea to provide comparisons using this minimal and fast neural network architecture to show that new regularizers can improve results for smaller CNNs.

\subsection{Use of more complex datasets}
\label{ss.complex_datasets}

The most common datasets used in regularization works concern objects and animals, which humans can easily distinguish. Another characteristic of these datasets is that they are perfectly balanced, meaning that every possible class has a similar amount of samples in the training and test validation. Usually, in medical and some real-world problems, such a balancing is hard to obtain.

In health-related problems, any increase in the results can lead to a safer treatment or even avoid misuse of medication and death. For these reasons, some datasets, like the Breast Cancer Histopathological Image Classification (BreakHis)~\cite{breakhis}, might be used to increase the work's relevance. In this specific case, where the results may infer in a life-threatening situation, the idea is to use a deeper CNN, like the Efficient-Net family~\cite{efficientnet} or ResNet~\cite{resnet}.

\subsection{Other problems besides classification}
\label{ss.other_problems}

In the past, CNNs and other neural networks were mainly used for the image classification task. However, more recently, CNNs were also employed in other tasks, such as object detection and speech recognition. For example, the YOLO architecture~\cite{yolo} is a  Fully Convolutional Network, which means that every layer performs a 2D convolutional process. In that sense, some changes on the loss calculation allow final layers to find out where objects on a given image are located. Another domain where CNNs have state-of-the-art results is image reconstruction. The Residual Dense Network has outstanding results on image reconstruction from noisy~\cite{rdn-reconstruction} and low-resolution images~\cite{rdn-sr}. 

There are two suggestions in this case. The first one is the use of regularization techniques in such different tasks or, at least, a reason for not using them in other domains. The second proposal is the development of new regularization targeting these specific problems. The only work found so far to solve different problems than image classification is the CutBlur~\cite{cutblur}, thus highlighting the lack of works in this direction.

\subsection{Source Code Links}
\label{ss.source_code_links}

As mentioned before, we only considered papers with the source code available. Table~\ref{t.sourcecode} presents the list of links concerning the source codes for every paper surveyed in this work.

\begin{table*}[htb!]
\centering

\begin{tabular}{lcll}
\multicolumn{1}{c}{Reference} & Short Name      & \multicolumn{1}{c}{Source Code} \\\hline
\cite{bag}                             & Bag of Tricks          & MXNet: \url{https://github.com/dmlc/gluon-cv} \\\hline
\cite{batchaugment}                             & Batch Augment          & PyTorch: \url{https://github.com/eladhoffer/convNet.pytorch}\\\hline
\cite{fixres}                             & FixRes          & PyTorch: \url{https://github.com/facebookresearch/FixRes}\\\hline
\cite{cutout}                             & Cutout          &  Pytorch: \url{https://github.com/uoguelph-mlrg/Cutout} \\\hline
\cite{cutmix}                             & CutMix          &  PyTorch: \url{https://github.com/clovaai/CutMix-PyTorch}\\\hline
\cite{randomerasing}                             & RandomErasing   & PyTorch: \url{https://github. com/zhunzhong07/Random-Erasing} \\\hline
\cite{mixup}                              & Mixup           &  PyTorch: \url{https://github.com/facebookresearch/mixup-cifar10}\\\hline
\cite{autoaugment}                        & AutoAugment     &  \begin{tabular}[c]{@{}l@{}}TensorFlow: https://github.com/tensorflow/tpu/\\blob/master/models/official/efficientnet/autoaugment.py \end{tabular}    \\\hline

\cite{fastautoaugment}                    & Fast AutoAugment     & PyTorch: \url{https://github.com/kakaobrain/fast-autoaugment} \\\hline
\cite{randaugment}                        & RandAugment    &  \begin{tabular}[c]{@{}l@{}}TensorFlow:  https://github.com/tensorflow/tpu/\\tree/master/models/official/efficientnet\end{tabular} \\\hline
\cite{pba}                             & PBA     & TensorFlow: \url{https://github.com/arcelien/pba}                     \\\hline

\cite{cutblur}                             & CutBlur         & PyTorch: \url{https://github.com/clovaai/cutblur} \\\hline
\cite{dropout}                             & Dropout         & \begin{tabular}[c]{@{}l@{}}PyTorch: https://pytorch.org/docs\\/stable/\_modules/torch/nn/modules/dropout.html  \end{tabular}              \\\hline
\cite{maxdropout}                          & MaxDropout      & Pytorch: \url{https://github.com/cfsantos/MaxDropout-torch/}                                                                           \\\hline
\cite{gradaug}                             & GradAug  & PyTorch: \url{https://github.com/taoyang1122/GradAug} \\\hline
\cite{localdrop}                             & Local Drop      & Available in the paper          \\\hline

\cite{shakeshake}                             & Shake-Shake     & Torch: \url{https://github.com/xgastaldi/shake-shake} \\\hline
\cite{shakedrop}                            & ShakeDrop       & Torch: \url{https://github.com/imenurok/ShakeDrop} \\\hline
\cite{manifoldmixup}                            & Manifold Mixup       & \begin{tabular}[c]{@{}l@{}}PyTorch: https://github.com/\\vikasverma1077/manifold\_mixup \end{tabular}  \\\hline
\cite{dropblock}                            & DropBlock       & \begin{tabular}[c]{@{}l@{}}TensorFlow: https://github.com/\\tensorflow/tpu/tree/master/models/official/resnet  \end{tabular}     \\\hline
\cite{autodropout}                            & AutoDrop       & \begin{tabular}[c]{@{}l@{}}TensorFlow: https://github.com/google-research/\\google- research/tree/master/auto\_dropout \end{tabular}                   \\\hline
\cite{inceptionv2v3}                            & Label Smoothing & TensorFlow: \url{https://github.com/tensorflow/models} \\\hline
\cite{tsla}                            & TSLA & Available in the paper \\\hline    
\cite{sls}                            & SLS & Available in the paper \\\hline  
\cite{jocor}                            & JoCoR & https://github.com/hongxin001/JoCoR\\\hline                                                          
\end{tabular}
\caption{Summarization of the approaches considered in the survey and their respective source codes.}
\label{t.sourcecode}
\end{table*}

\section{Conclusion}
\label{s.conclusion}

Regularization is a vital tool to improve the final CNN results since it helps prevent the model from overfitting on the training data. This work aimed at showing the most recent commitments in the area, targeting to deliver a brief resume on how they work and their main results. 

\review{This work introduced a lineup of recent regularizers that can fit in most neural networks for outcome improvement. Although some can drastically increase the training time, such as AutoAugment, most do not require any relevant extra time, and none influences the time taken for inference. Right after the introduction, we provide a brief explanation of how CNN works and a little history of its development, and then we divided all works analyzed in this paper as follows:}

\begin{itemize}
\item "input regularization", where the models work before the image is fed to the network;
\item "internal regularization", when the regularization algorithms work after the image is feedforwarded to the model; and
\item "label regularization", when the algorithm performs on the output layer.
\end{itemize}
\review{Besides, the methodology presents the most popular datasets used to evaluate regularization techniques and the most traditional CNN architectures for such a task. Such information is crucial, for it helps standardize an evaluation protocol from now on.}

\review{Along with the reported results for each work, we provided our opinion on setting up a state-of-the-art regularizer, an essential but trustful protocol evaluation for new regularizers, which can help compare the results and provide insights for researchers in this area. The same section highlights some issues we found in most of the works:}

\begin{itemize}
  \item \review{the lack of using simpler architectures, which are the ones that could be more benefited from the use of regularizers;} and
  \item \review{the lack of an evaluation of methods on more complex data, such as unbalanced data sets, to provide richer information for other researchers.}
\end{itemize}
\review{Last but not least, we encourage the development of new regularization techniques on tasks other than image classification, such as object detection and image reconstruction.}

\begin{acks}
The authors are grateful to the S\~ao Paulo Research Foundation through grants 2014/12236-1, 2017/25908-6, and 2019/07665-4, as well as the Brazilian National Council for Scientific, Technological Development grants 307066/2017-7 and 427968/2018-6 and Eldorado Research Institute.
\end{acks}

%
\bibliographystyle{ACM-Reference-Format}
\bibliography{sections/refs}

\end{document}